%% file: main.tex
\documentclass[sigconf]{acmart}

\usepackage[utf8]{inputenc} % allow utf-8 input
\usepackage[T1]{fontenc}
\usepackage{hyperref}
\usepackage{url}
\usepackage{amsfonts}
\usepackage{nicefrac}
\usepackage{microtype}
\usepackage{graphicx}
\usepackage{natbib}
\usepackage{listings}
\usepackage[longtable]{multirow}

\usepackage{color}

\definecolor{bluekeywords}{rgb}{0.13, 0.13, 1}
\definecolor{greencomments}{rgb}{0, 0.5, 0}
\definecolor{redstrings}{rgb}{0.9, 0, 0}
\definecolor{graynumbers}{rgb}{0.5, 0.5, 0.5}

\copyrightyear{2023}
\acmYear{2023}
\setcopyright{acmlicensed}
\acmConference[SIGIR '23]{Proceedings of the 46th International ACM SIGIR Conference on Research and Development in Information Retrieval}{July 23--27, 2023}{Taipei, Taiwan}
\acmBooktitle{Proceedings of the 46th International ACM SIGIR Conference on Research and Development in Information Retrieval (SIGIR '23), July 23--27, 2023, Taipei, Taiwan}
\acmPrice{15.00}
\acmDOI{10.1145/3539618.3591892}
\acmISBN{978-1-4503-9408-6/23/07}

\begin{document}
\sloppy

\title[\texttt{tieval}: An Evaluation Framework for Temporal Information Extraction Systems]{\texttt{tieval}: An Evaluation Framework for \\ Temporal Information Extraction Systems}

\author{Hugo Sousa}
\orcid{0000-0003-3226-9189}
\affiliation{%
	\institution{INESC TEC \and University of Porto}
	\city{Porto}
	\country{Portugal}
}
\email{hugo.o.sousa@inesctec.pt}

\author{Ricardo Campos}
\orcid{0000-0002-8767-8126}
\affiliation{%
	\institution{INESC TEC}
	\city{Porto}
	\country{Portugal}
}
\affiliation{%
	\institution{Polytechnic Institute of Tomar \and Ci2 - Smart Cities Research Center}
	\city{Tomar}
	\country{Portugal}
}
\email{ricardo.campos@ipt.pt}

\author{Alípio Jorge}
\orcid{0000-0002-5475-1382}
\affiliation{%
	\institution{INESC TEC \and University of Porto}
	\city{Porto}
	\country{Portugal}
}
\email{amjorge@fc.up.pt}

\begin{abstract}

	Temporal information extraction (TIE) has attracted a great deal of interest over the last two decades. Such endeavors have led to the development of a significant number of datasets. Despite its benefits, having access to a large volume of corpora makes it difficult to benchmark TIE systems. On the one hand, different datasets have different annotation schemes, which hinders the comparison between competitors across different corpora. On the other hand, the fact that each corpus is disseminated in a different format requires a considerable engineering effort for a researcher/practitioner to develop parsers for all of them. These constraints force researchers to select a limited amount of datasets to evaluate their systems which consequently limits the comparability of the systems. Yet another obstacle to the comparability of TIE systems is the evaluation metric employed. While most research works adopt traditional metrics such as precision, recall, and $F_1$, a few others prefer temporal awareness -- a metric tailored to be more comprehensive on the evaluation of temporal systems. Although the reason for the absence of temporal awareness in the evaluation of most systems is not clear, one of the factors that certainly weighs on this decision is the need to implement the temporal closure algorithm, which is neither straightforward to implement nor easily available. All in all, these problems have limited the fair comparison between approaches and consequently, the development of TIE systems. To mitigate these problems, we have developed \texttt{tieval}, a Python library that provides a concise interface for importing different corpora and is equipped with domain-specific operations that facilitate system evaluation. In this paper, we present the first public release of \texttt{tieval} and highlight its most relevant features. The library is available as open source, under MIT License, at PyPI\footnote{\url{https://pypi.org/project/tieval/}} and GitHub\footnote{\url{https://github.com/LIAAD/tieval}}.

\end{abstract}




\maketitle

\begin{figure}[ht]
	\includegraphics[width=\linewidth]{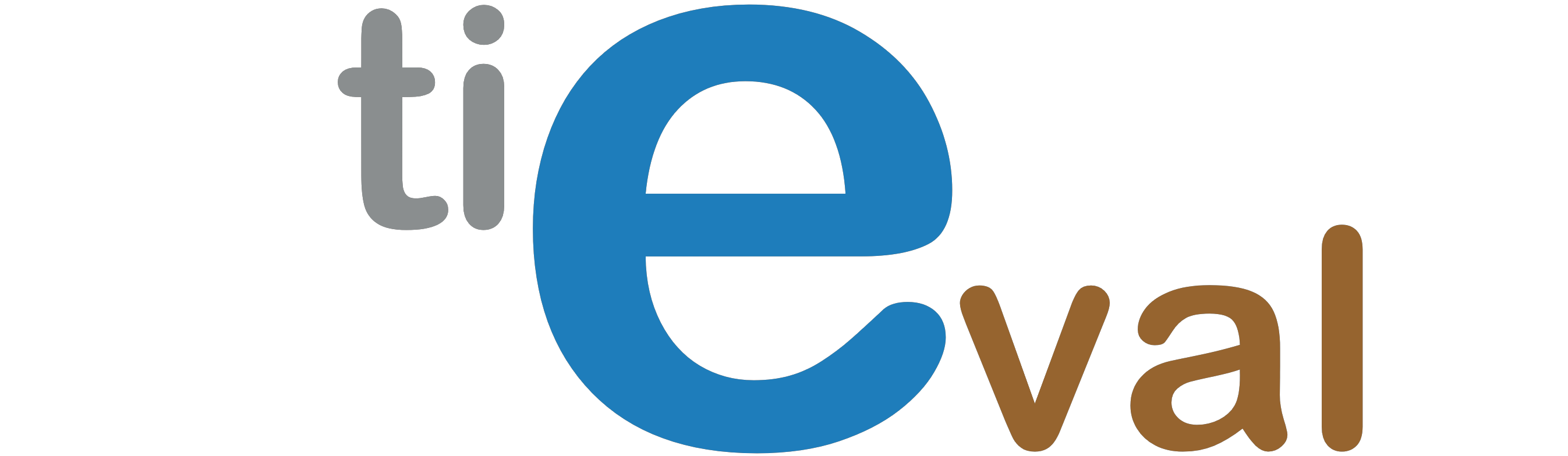}
	\Description{\texttt{tieval} logo.}
	\caption{\texttt{tieval} logo.}
	\label{fig:logo}
\end{figure}

\section{Introduction}

\begin{figure}[ht]
	\centering
	\includegraphics[width=\linewidth]{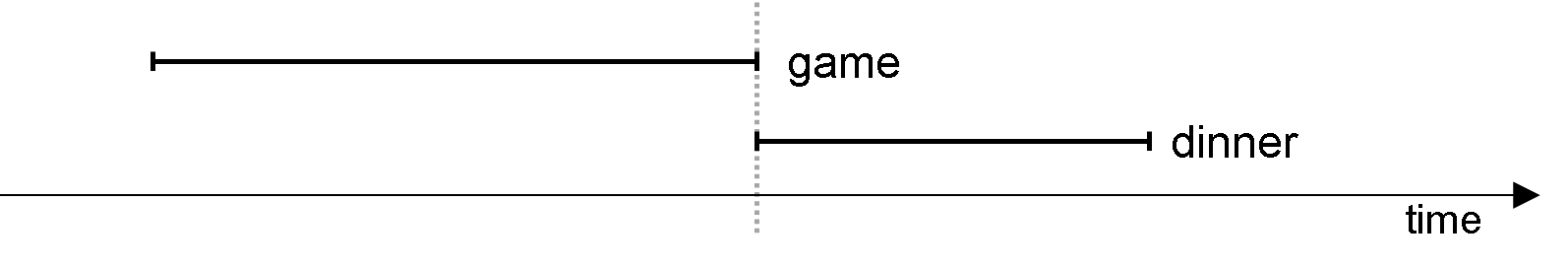}
	\Description{Relative timeline of events that can be inferred from the running example.}
	\caption{Relative timeline of events that can be inferred from the running example.}
	\label{fig:timeline}
\end{figure}

Understanding the temporal order of events is essential to human communication. We, humans, can easily understand the relative order of events in a conversation or when reading a news article. However, many challenges are raised when we try to automate such tasks with a computer program, or in a more general setting of understanding the underlying narrative~\cite{Santana2023AData}. The first difficulty that emerges is how to represent temporal information. Since in most cases, we do not explicitly specify the start and end time of each event, temporal information, such as order and time span, ends up being inferred from the events themselves. In this regard, computer algorithms can make use of temporal clues in the text, and of external sources, such as knowledge bases, to anchor events on a timeline. For instance, in the sentence ``We went to dinner after the game.'', two events, ``dinner'' and ``game'', can -- supported on the word ``after'' -- be identified and used to recreate a timeline of events, despite the lack of explicit temporal information (see Figure~\ref{fig:timeline}). The ordering of events and the knowledge about them can be further expanded if used together with appropriate external sources. For instance, the event ``game'' can be contextualized and anchored on the timeline by searching for information on a knowledge base. However, in the case of the ``dinner'' event, it turns out impossible to know the exact time of occurrence unless it is specified in the text. This shows that representing temporal information is not a trivial task, since there are several borderline cases for which no standard approach has been established.

Over the years, and particularly in the last two decades, this problem has been highly studied, leading to several proposals from the research community~\cite{Campos2014SurveyApplications,Leeuwenberg2019AText}. These proposals led to the numerous annotation schemes and corpora that we have today at our disposal~\cite{Naik2019TDDiscourse:Events,Ning2018ARelations,UzZaman2013SemEval-2013Relations}. Although these efforts have been essential to mature temporal information extraction and its subtasks -- such as temporal expression identification or temporal relation classification -- they also pose some problems in the process of benchmarking different methods. One of the problems has its roots in the fact that evaluating the methods, often requires reading multiple corpora, each of which has a different perspective on temporal representation, ultimately preventing comparability among the different methods and corpora. This is compounded by the fact that corpora are stored in a variety of formats (e.g., XML, TimeML, or table), which requires a considerable engineering effort to load them all.

Another issue that limits the comparison between systems is the lack of standardization in the metrics used in the evaluation process. This is a particular problem of temporal relation extraction -- a subtask of TIE, which deals with the identification and classification of the temporal relations between entities -- where different metrics are often employed during the evaluation process. While, initially, systems were evaluated and compared using standard metrics, such as recall, precision, and F-score~\cite{Verhagen2007SemEval-2007Identification,Verhagen2010SemEval-2010TempEval-2}, more recently, metrics such as temporal awareness~\cite{UzZaman2011TemporalEvaluation} have proven to be more reliable in the evaluation of temporal relation extraction methods. The reasoning behind this is that, while traditional metrics focus on the local effectiveness of the model, temporal awareness better understands the relative order of events by considering the global temporal structure of the predictions. This is accomplished by taking into account the temporal relations that can be inferred from the established ones (a process typically referred to as temporal closure), making this a more comprehensive metric for evaluating temporal systems. Despite the emergence of this temporal awareness, many studies still rely solely on traditional metrics to evaluate their system. We speculate that this is due to the fact that temporal awareness requires domain-specific operations such as temporal closure -- which are not (yet) readily available in any framework and therefore require individual implementation by each research group. In addition, temporal awareness requires the implementation of a strategy to deal with inconsistent predictions of the system, which is generally not explored in recent studies.

To mitigate the above-referred issues, we developed \texttt{tieval} (logo presented in Figure~\ref{fig:logo}), a Python library that enables the development and evaluation of TIE systems. This framework provides a simple interface to download and read TIE corpora in various formats. It currently covers well-known corpus -- such as TempEval-3~\cite{UzZaman2013SemEval-2013Relations}, TDDiscource~\cite{Naik2019TDDiscourse:Events}, and MeanTime~\cite{Minard2016MEANTIMECorpus} -- and lays the foundations for others to be included by providing base classes for the construction of the corpus. It also provides domain-specific operations -- such as temporal closure and simple translation of intervals into point relations -- that can be used to develop TIE systems. Besides that, it includes an evaluation infrastructure for a comprehensive assessment of the effectiveness of the different models. Because \texttt{tieval} supports the entire development pipeline of TIE, it can also be used to ensure reproducibility and fair benchmarking of future research. The main contributions of \texttt{tieval} are the following:

\begin{enumerate}
	\item it gathers and standardizes the multiple corpora for the development of TIE
	      systems, making it easy to access with just a few lines of code;
	\item it facilitates access to domain-specific operations, such as temporal closure,
	      and metrics, such as temporal awareness;
	\item it provides a standard framework, thus making it easy for new methods to be
	      compared against previous ones.
\end{enumerate}

The remainder of the paper is organized as follows: the next section provides an overview of recent work in TIE and some of its software. We then proceed to present the \texttt{tieval} package in section~\ref{sec:tieval}. This section starts with a general introduction and then goes into detail about some of its most relevant features. Section~\ref{sec:future_work} serves to present our thoughts on what we strive to be the next steps in the development of the framework.

\section{Related Work}
\label{sec:related_work}

\subsection{Temporal Information Extraction Overview}
Extracting temporal information from documents written in natural language in an inter-operable format has long been an interest of the artificial intelligence community~\cite{Ling2010TemporalExtraction,Derczynski2015TimeIssue}. Since the introduction of the Time Markup Language (TimeML)~\cite{Pustejovsky2003TimeML:Text.}, in 2003, the temporal graph has become the de-facto standard to represent temporal information. In this graph, the nodes are temporal entities and the edges are the temporal relation that holds between them. The temporal entities can take two forms: event expressions, which are defined as situations that happened (e.g., ``went'' or ``bought''); and temporal expressions (timex), which can convey temporal information explicitly (e.g., ``October 27, 1996'') or implicitly (e.g., ``a few years ago'')~\cite{Campos2017IdentifyingQueries}. The temporal relations are held in the form of temporal links (tlink) that contain temporal relations between pairs of events (E-E relations), events and time expressions (E-T relations), and events and document creation time (E-DCT relations), where DCT is a special timex that stores the document creation time. Overall, these temporal relations can take thirteen types, which is the number of relations that can exist between two-time intervals~\cite{Allen1983MaintainingIntervals}.

The first corpus that was annotated with this scheme was TimeBank~\cite{Pustejovsky2003TheCorpus}. The release of this corpus, also dated from 2003, sparked a wave of research in the field. Most notably TimeBank was used in the first edition of the TempEval shared tasks~\cite{Verhagen2007SemEval-2007Identification} in 2007, which ended up having a sequel in 2010~\cite{Verhagen2010SemEval-2010TempEval-2} and 2013~\cite{UzZaman2013SemEval-2013Relations}. One of the outcomes of these shared tasks was the partitioning of TIE into a set of sub-problems which can be conceptually defined as temporal entity identification, tlink identification, and tlink classification. Although some works developed systems for more than one of these sub-tasks, most of the systems are concerned with only one of them. Furthermore, temporal entity identification systems are traditionally partitioned into subsystems for several classes of temporal entities. For example, for the TimeBank corpus, one system is usually trained to identify events and another to identify timexs. The \texttt{tieval} architecture follows this partition of TIE.

\subsection{Corpora}
The TimeBank corpus, and more abstractly, the TimeML annotation scheme was widely studied by the community. Such scrutiny led to the emergence of several new corpora. Some used the TimeML annotation scheme to annotate different corpora, such as AQUAINT~\cite{Graff2002TheLDC2002T31} and the Platinum corpus~\cite{UzZaman2013SemEval-2013Relations}, while others were concerned about extending the annotation scheme to other languages. The most remarkable effort on this domain was the TempEval-2 shared task~\cite{Verhagen2010SemEval-2010TempEval-2} that produced corpora for Chinese~\cite{Li2014ChineseHeidelTime}, French~\cite{Bittar2011FrenchCorpus}, Italian~\cite{Caselli2011AnnotatingIta-TimeBank}, and the Spanish~\cite{GuerreroNieto2011ModeSCorpus} language. Another noteworthy effort is the MeanTime corpus~\cite{Minard2016MEANTIMECorpus} in which the authors annotated $120$ news articles written in English from Wikinews\footnote{\url{https://en.wikinews.org/}}, and translated the texts into Italian, Spanish, and Dutch. Costa and Branco~\cite{Costa2012TimeBankPT:Portuguese} followed a similar process to construct TimeBankPT, translating the original TimeBank to European Portuguese and manually adjusting the annotations when needed.

Apart from the extensions to other languages, the TimeML annotation scheme was also extended to other domains. A concrete example is the case of the clinical domain for which two corpora have been produced, the i2b2~\cite{Sun2013EvaluatingChallenge} and THYME~\cite{Styler2014TemporalDomain} datasets\footnote{These corpora are not available for open access and, as a consequence, we were not able to include them on the framework.}. Further significant contributions were the proposals that explored ways to mitigate some of the issues found on the TimeBank annotation effort, such as: sparse annotation, as is the case of TimeBank-Dense~\cite{Cassidy2014AnOrdering} and TDDiscourse~\cite{Naik2019TDDiscourse:Events}; improve inter-annotator agreement, as in the MATRES~\cite{Ning2018ARelations} dataset; and include other sources of knowledge, as the  TCR~\cite{Ning2018JointRelations} and RED~\cite{OGorman2016RicherAnnotation} datasets.

Aside from the TimeML, and related approaches, there have also been other proposals that were explored by the research community. One of them is absolute timeline placement, in which the temporal entities are directly anchored on a timeline by labeling each entity with the time (or time span) of occurrence. The most remarkable efforts in this direction were produced by Reimers et al.~\cite{Reimers2016TemporalCorpus} -- which produced the EventTime corpus by annotating the events in TimeBank with a specific day, or span of days -- and Leeuwenberg and Moens~\cite{Leeuwenberg2020TowardsReports} -- which annotated $169$ clinical records from the i2b2 corpus with the most likely start and end time of each event along with a lower and upper bound.

This shows that several corpora have been introduced for the TIE task. However, the fact that they were released in different formats makes it hard to leverage their power, which is one of the issues mitigated by \texttt{tieval}.

\subsection{Software}
To the best of our knowledge, the only framework that made available TIE operations -- including temporal closure and temporal awareness -- is the Anafora Tools project\footnote{\url{https://github.com/bethard/anaforatools}} which was built to work with files stored in the Anafora XML format~\cite{Chen2013Anafora:Tool}, used to annotate the THYME corpus~\cite{Styler2014TemporalDomain}. The framework presented in this paper aims to be a more general tool, unifying all corpora in a single format.

\section{tieval}
\label{sec:tieval}
The vision for \texttt{tieval} was to build a framework that would support and facilitate the evaluation of TIE systems. With the development of libraries such as Numpy, TensorFlow, and PyTorch, Python has established itself as the programming language of choice within the machine learning community. For that reason, \texttt{tieval} was built in Python. To facilitate the installation we made it available on Python Package Index (PyPI)\footnote{\url{https://pypi.org/project/tieval/}}. Thus, the toolkit can be easily installed through \texttt{pip}, as follows:
\begin{verbatim}
	$ pip install tieval==0.0.5
\end{verbatim}

In this paper, we will use version \texttt{0.0.5}, which is the first and the most recent version of the package. However, the reader is advised to install the newest release at the time of reading the paper and refer to the project repository for up-to-date documentation. Furthermore, for users that might be interested in contributing to \texttt{tieval}, we encourage forking the source repository and making a pull request.

\texttt{tieval} contains three modules that represent the three cornerstones of any machine learning project: \texttt{datasets}, \texttt{models}, and \texttt{evaluation}. The \texttt{datasets} module is responsible for downloading and reading the corpora available for TIE, the \texttt{models} module is responsible for the construction of the models, and the \texttt{evaluation} module has methods to make a proper evaluation for each of the TIE tasks. In the following sections, we will present the inner workings of the framework with scripts to exemplify its usability.

\subsection{Datasets}

\begin{figure}[htbp]
	\centering
	\includegraphics[width=\linewidth]{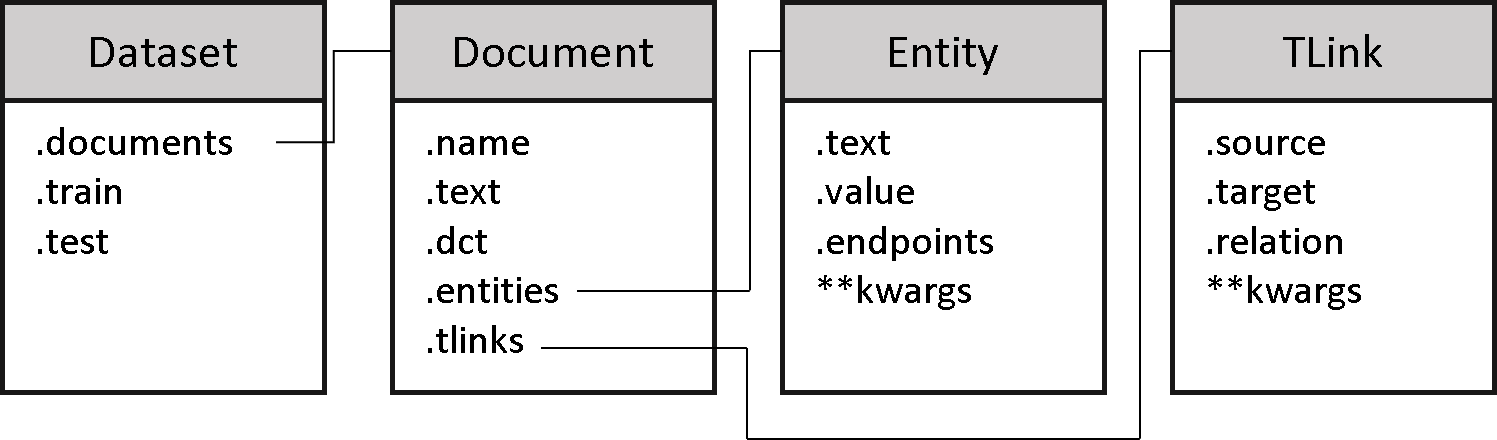}
	\Description{Objects used to represent a dataset on \texttt{tieval}. The arrow represents a relation of ``Iterable''.}
	\caption{Objects used to represent a dataset on \texttt{tieval}. The arrow represents a relation of ``Iterable''.}
	\label{fig:dataset}
\end{figure}

With \texttt{tieval}, we intend to mitigate the above-referred issues by making it easy for the user to work with several corpora with a few lines of code. To that end, we developed an architecture that would unify the different annotations and storing formats of the corpus. This architecture is composed of several objects which are depicted in Figure~\ref{fig:dataset}.

The \texttt{Dataset} object is the final representation of each corpus. It compiles the set of all the documents in the corpus on the \texttt{documents} attribute which is segmented into the train and test attributes whenever provided in the original paper\footnote{When no standard train/test split is provided by the authors all the documents are placed on the \texttt{train} attribute.}. Each document is then stored as an instance of the \texttt{Document} class (see the Document grey box in Figure~\ref{fig:dataset}), which contains all the information necessary for TIE, more
specifically:

\begin{description}
	\item[name] a string that contains the name of the document (e.g. ``wsj\_0026.tml'');
	\item[text] a string with the raw text of the document;
	\item[dct] is a \texttt{Timex} that contains the document creation time (e.g. \texttt{Timex("12-10-2004")});
	\item[entities] is the set of Entities -- either a \texttt{Timex} or \texttt{Event} -- that are annotated on the corpus. Each \texttt{Entity} is, at its core, a data class made to store all the info provided on the annotation. Therefore, it has to be flexible to accommodate the different types of information provided in the different corpus. For instance, the GraphEve corpus provides the lemma for each event while TempEval-2 does not;
	\item[tlinks] a set o \texttt{TLink}'s that stores the temporal relations annotated on the document. Each \texttt{TLink} contains a \texttt{source} and \texttt{target} entity as well as the temporal relation between them -- on the \texttt{relation} attribute.
\end{description}

A special remark needs to be made about the \texttt{relation} attribute of the \texttt{TLink} object. When creating a \texttt{TLink} instance, one must specify the temporal relation that exists between the two temporal entities, referred to as the source and target entities. In the majority of corpora, this typically involves one of the thirteen temporal relations proposed by Allen~\cite{Allen1983MaintainingIntervals}, which lists the possible relationships between two-time intervals. However, certain corpora, such as TempEval-2 and MATRES, exhibit more flexibility in terms of relation types.

\begin{figure}[htbp]
	\centering
	\includegraphics[width=\linewidth]{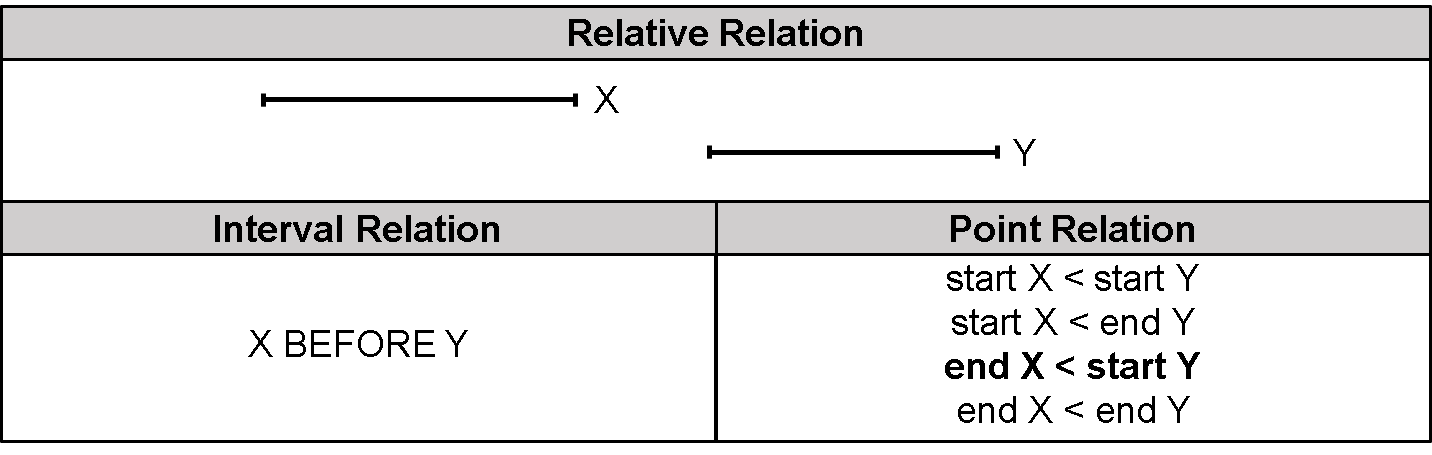}
	\Description{How the interval relation {\small BEFORE} is represented as a point relation.}
	\caption{How the interval relation {\small BEFORE} is represented as a point relation.}
	\label{fig:relation_table}
\end{figure}

In the case of TempEval-2, annotators were permitted to use more ambiguous relations, such as {\small BEFORE-OR-OVERLAP} and {\small OVERLAP-OR-AFTER}. For MATRES, annotators were instructed to provide the temporal relation between the starting points of the temporal entities. To accommodate these variations in annotation types, we have developed a \texttt{TemporalRelation} object that manages the annotated relation. Within this object, every relation is represented using point relations, as opposed to the traditional interval relations. Figure~\ref{fig:relation_table} illustrates the conversion of the interval relation {\small BEFORE} into a point relation and includes a relative relation for demonstration purposes.

By implementing the \texttt{TemporalRelation} object, we can effectively handle the diverse types of annotated temporal relations present in different corpora. This approach ensures that our architecture remains adaptable and inclusive, allowing for seamless integration and analysis of various datasets in temporal information extraction research.

Note that the {\small BEFORE-OR-OVERLAP} relation on TempEval-2 represents an uncertainty of the annotator between the end time of the source entity and the start time of the target entity, however, the annotator is certain about the remaining point relations. Further note that, although we explicitly state four-point relations in Figure~\ref{fig:relation_table}, upon the adaptation of the current datasets into \texttt{tieval} format, three of them are redundant, as the point relation ``end A < start B'' completely defines the remaining point relations. Therefore, on \texttt{tieval}, whenever there is a new dataset to include, the user can provide the relation in the way that is most appropriate, as shown in Listing~\ref{lst:tlinks}.

\begin{lstlisting}[
	caption={Different ways to pass the temporal relation to the \texttt{TLink} object. The first argument (X) is the source entity, the second (Y) is the target entity, and the third is the temporal relation between them. This can be passed as an interval relation, {\small BEFORE}, or as a point relation, in the form of a dictionary structure. On the latter, the interpretation for the expected keys is the following: ``x'' and ``y'' stands for the source and target entity, respectively; while the ``s'' and ``e'' stand for ``start'' and ``end''. As an example, ``xe\_ys'' is the point relation between the source end and the target start.},
	label={lst:tlinks},
	language=Python,
]
from tieval.links import TLink
tl1 = TLink("X", "Y", "before")
tl2 = TLink("X", "Y", {"xe_ys": "<"})
tl3 = TLink("X", "Y", {
    "xs_ys": "<", 
    "xs_ye": "<", 
    "xe_ys": "<", 
    "xe_ye": "<"
})
\end{lstlisting}

In order to establish a consistent representation across the various corpora, we have designed a unique reader for each corpus. These specialized dataset readers inherit from an abstract base class called \texttt{BaseDocumentReader}, which mandates the implementation of five methods corresponding to the five attributes used to create a \texttt{Document} instance: name, text, dct, entities, and tlinks.

To extract the required information, the base class incorporates three attributes: the path of the document being read; the content of the dictionary generated by parsing the document using the \texttt{xmltodict}\footnote{\url{https://pypi.org/project/xmltodict/}} library; and the \texttt{xml} attribute, which is obtained by parsing the file with the \texttt{xml}\footnote{\url{https://docs.python.org/3/library/xml.etree.elementtree.html}} library. It is worth mentioning that although \texttt{json} has become the standard format for information exchange in recent years, we opted for \texttt{xml} since the majority of the datasets were stored in this format.

An example of how this approach is implemented can be seen in the script provided in Listing~\ref{lst:read_document}, which demonstrates how to read a document from the TempEval-3 corpus using the \texttt{TempEval3DocumentReader}.

\begin{lstlisting}[
	caption={Read a document of the TempEval-3 corpus.},
	label={lst:read_document},
	language=Python
]
from tieval import datasets
path = "tempeval-3/wsj_0026.tml"
reader = datasets.TempEval3DocumentReader(path)
doc = reader.read()
\end{lstlisting}

To fully integrate a new corpus on the library -- and automatically read the entire corpus -- the user just needs to add an entry on the \texttt{DATASETS\_METADATA} dictionary with the metadata necessary to read the document. This information will be used on the \texttt{read} function of the \texttt{datasets} module, which only requires the name of the corpus to produce an instance of the \texttt{Dataset} object with all the annotations provided in there. The script in Listing~\ref{lst:read} presents how to perform such an operation.

\begin{lstlisting}[caption={Read the TempEval-3 corpus.}, label={lst:read}, language=Python]
from tieval import datasets
te3 = datasets.read("TempEval_3")
\end{lstlisting}

The current version of \texttt{tieval} natively supports the download and reading of an extensive list of corpora for TIE. A complete list of the corpora considered is provided in Table~\ref{tab:corpus}. In order to ensure long-term support for these corpora, we created a repository with them. Besides that, it also has the advantage that we can standardize the structure of the folders and add useful information to the raw datasets (for instance, the spans of the temporal entities identified on the text) and fix errors on the original annotation\footnote{The changes made on the original corpus are detailed on the file \texttt{logbook.rst} in the \texttt{docs} folder of the GitHub project repository \url{https://github.com/LIAAD/tieval}.}. For that reason, we were careful to verify the license for each of the corpora and publish only the ones that allowed for redistribution or did not provide any license.

\begin{table*}[ht]
	\centering
	\caption{The corpora currently supported on \texttt{tieval}.\label{tab:data}}
	\label{tab:corpus}
	\include{tables/data}
\end{table*}

\subsection{Models}
The current version of \texttt{tieval} has four built-in models, namely: a baseline for timex identification; the HeidelTime model~\cite{Strotgen2013HeidelTime:TempEval-3} for timex identification and classification; a baseline for event identification; and the CogCompTime 2.0 model~\cite{Ning2019AnExtraction} for tlink classification. The availability of these four models is intended for practitioners that may want to experiment using any layer of temporal information in their specific application. Apart from that, it also provides researchers with the implementation of baseline models for reference in their work.

For the baseline models, we provide pre-trained weights, however, the user can also train the model from scratch. A description of each of the models is provided below:

\begin{description}
	\item[TimexIdentificationBaseline] This baseline is the spaCy\footnote{\url{https://spacy.io/}} named entity recognition model trained (from scratch) on the train documents of the TempEval-3 corpus to identify timexs.
	\item[EventIdentificationBaseline] This model has the same architecture of the \texttt{TimexIdentificationBaseline} but was trained to identify events rather than timexs on the TempEval-3 corpus.
	\item[HeidelTime] This model is a widely recognized multilingual temporal tagger that was originally written in Java\footnote{\url{https://github.com/HeidelTime/heideltime}}. However, there have been efforts to build python wrappers. In \texttt{tieval} we used the \texttt{py\_heideltime} wrapper which is available on GitHub\footnote{\url{https://github.com/hmosousa/py_heideltime}}.
	\item[CogCompTime2] This model leverages the ELMo~\cite{Peters2018DeepRepresentations} word embeddings and the TempProb~\cite{Ning2018ImprovingResource} knowledge base to classify the temporal relation between a pair of temporal entities~\cite{Ning2019AnExtraction}. Our implementation was adapted from the repository made available\footnote{\url{https://github.com/qiangning/NeuralTemporalRelation-EMNLP19}} by the authors.
\end{description}

Listing~\ref{lst:models} presents a script that would download the baseline model for temporal expression identification (\texttt{TimexIdentificationBaseline}), train the model on the TempEval-3 train set, and produce predictions for the TempEval-3 test set.

\begin{lstlisting}[caption={How to download, train, and predict with for the temporal identification task.}, label={lst:models}, language=Python]
from tieval import models
model = model.TimexIdentificationBaseline()
model.fit(te3.train)
predictions = model.predict(te3.test)
\end{lstlisting}

A user interested in releasing his/her model in \texttt{tieval} can do it by creating a subclass of one of our base classes for models. There are two base classes: a \texttt{BaseModel} which just requires the implementation of the \texttt{predict} method which is intended for models that are available in other repositories -- for instance, the HeidelTime model -- and a \texttt{BaseTrainableModel} which, besides the \texttt{predict}, requires the implementation of the \texttt{fit} method, which implements the training loop for the model.

\subsection{Evaluation}
\label{ssec:evaluation}

\begin{table*}[ht]
	\centering
	\caption{The results obtained by evaluating the four models integrated in \texttt{tieval} on the Platinum (TempEval-3 test set), TCR, and MeanTime (the English version) corpus. P stands for precision, R for recall, F1 is the F1-score, and TF1 is the temporal awareness. All the results in the table are micro metrics.}
	\label{tab:results}
	\include{tables/results}
\end{table*}

\texttt{tieval} provides an evaluation function for four subtasks of TIE, more specifically: timex identification, event identification, tlink identification, and tlink classification.

The input is standard for all the evaluation functions: \texttt{annotations}, a dictionary with the name of the documents as keys and the annotations as values; \texttt{predictions}, follows the same structure as \texttt{annotations} but for each document, key contains the predictions made by a model. The output of the functions is dependent on the task being evaluated. For the identification tasks (timex, event, and tlink) the function produces the standard macro and micro metrics for precision, recall, and $F_1$-score. Listing~\ref{lst:evaluate} presents a script that evaluates the predictions made by the event baseline model in the TempEval-3 test set.

\begin{lstlisting}[caption={Evaluate event baseline model on the TempEval-3 test set.}, label={lst:evaluate}, language=Python]
from tieval import evaluate
annotations = {
    doc.name: doc.events 
    for doc in te3.test
}
result = evaluate.event_identification(
    annotations, 
    predictions
)
\end{lstlisting}

Table~\ref{tab:results} depicts the results obtained by the implemented models on four benchmark corpus. Note that $TF_1$ is the temporal awareness metric and is only computed for \texttt{CogCompTime2} (the only tlink classification system). Another interesting remark is the fact that the \texttt{TimexIdentificationBaseline} achieves effectiveness comparable to \texttt{HeidelTime} despite its simplicity.

The tlink classification is the most elaborate evaluator as it also computes the temporal awareness metric~\cite{UzZaman2011TemporalEvaluation}. The complexity of the calculation of temporal awareness lies in the computation of temporal closure. With \texttt{temporal\_closure} the closure operation can be easily performed on the document level, with the \texttt{closure} method of the \texttt{Document} object, or applied to a set of tlink's with the \texttt{temporal\_closure} function available on the library. The script in Listing~\ref{lst:closure} illustrates how to perform such operations.

\begin{lstlisting}[
	caption={How to compute the temporal closure with a \texttt{Document} object and with a set of \texttt{TLink}'s.},
	label={lst:closure},
	language=Python
]
from tieval import temporal_closure
doc = te3["wsj_0026.tml"]
closure_tlinks = doc.closure
closure_tlinks = temporal_closure(doc.tlinks)
\end{lstlisting}

For the temporal closure to be efficiently performed, on the backend, the closure operation is executed with a point-based reasoner which was inspired by the work of Gerevini et al.~\cite{Gerevini1993TemporalIII}. As stated above, each \texttt{TLink} instance contains an attribute named \texttt{relation} which is an instance of the \texttt{TemporalRelation} object. Within the \texttt{TemporalRelation} all temporal relations are represented as the point relations by the means of a \texttt{PointRelation} instance. In the point representation, there are only four types of temporal relations, namely before (<), after (>), equal (=), and not defined (None). With this point relation one can build a directed graph (henceforth referred to as timegraph) where the nodes are the entities' endpoints (start and end of the entity) and the edges represent the before (<) relation. This is accomplished by reflecting the after (>) relations and aggregating the equal (=) relations in a single node.

\begin{figure}[ht]
	\centering
	\includegraphics[width=\linewidth]{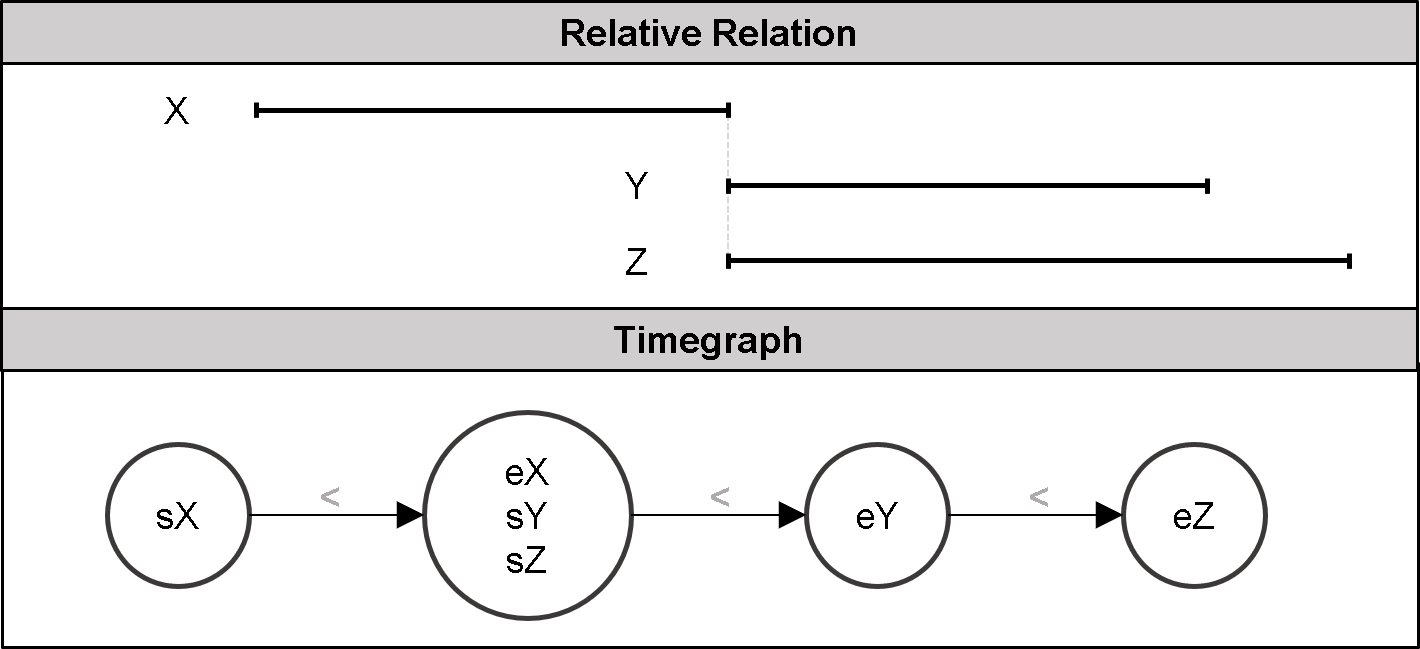}
	\Description{On the top part of the image is the relative relations between entities X, Y, and Z. On the bottom is the graphical representation of the timegraph that would be generated.}
	\caption{On the top part of the image is the relative relations between entities X, Y, and Z. On the bottom is the graphical representation of the timegraph that would be generated.}
	\label{fig:timegraph}
\end{figure}

In the timegraph, inferring temporal relations is reduced to the problem of finding if two entities' endpoints are connected, i.e., they are in the same subgraph (by subgraph we mean a fully connected graph of the timegraph). If that is the case, one can retrieve the endpoints on the entity pair and validate if the order of the entity endpoints is a valid temporal relation. To clarify this concept, Figure~\ref{fig:timegraph} presents the timegraph built for a scenario where two tlinks were provided: X {\small MEETS} Y and Y {\small STARTS} Z. To infer the temporal relation between X and Z one must query the endpoints in the timegraph. In this case, one would get the following sequence of endpoints: sX < ex = sZ < eZ. After retrieving the sequence of endpoints one just needs to validate if that sequence is a valid interval relation. In this example, one can conclude that the temporal relation between X and Z is MEETS.

To get a practical understanding of the runtime of the temporal closure algorithm, we executed it on all documents currently available in\texttt{tieval}. On a computer with an Intel Core i5-8500 CPU, the algorithm took less than half a second for $95\%$ of the documents, while the worst-case scenario took roughly $1.6$ seconds.

This finalizes the presentation of the main functionalities, and some inner workings, of the first version of \texttt{tieval}. The current version already provides functionalities that (we believe) will be beneficial for the TIE community. However, we already have some ideas to further improve this library. These ideas are discussed in section~\ref{sec:future_work}.

\subsection{Observations}
While building \texttt{tieval}, and in particular the \texttt{datasets} module, we found some inconsistencies in the corpus we were working with. For instance, we found that the articles \texttt{APW20000115.0209} and \texttt{APW20000107.0088} of the AQUAINT corpus contained the same news article, differing only in the annotations and in the value of the document creation time. These inconsistencies were mitigated by implementing data-cleaning processes that changed the original annotations. Consequently, the results on the \texttt{tieval} framework will (most frequently) not resemble the exact result that was reported in previous works, even if the same model is employed.

\section{Conclusion and Future Work}
\label{sec:future_work}

This paper introduces the first public release of the \texttt{tieval} package, an open-source Python library for the development and evaluation of TIE systems. \texttt{tieval} provides several functionalities to facilitate research in this field. These include importing multiple benchmark corpora in different formats, domain-specific operations such as temporal closure or transformation from interval to point relations, out-of-the-box baseline systems, and evaluation measures for TIE tasks. Therefore, it provides the community with a standard way to benchmark TIE systems in a fair and comparable way, while enabling the development of reproducible systems.

For future versions of the package, we aim to extend its functionalities. One idea we are keen to implement is visualization techniques to display the relative timeline of events from the annotations. In addition, we will add methods to include other levels of information when available such as coreference resolution in the MeanTime corpus~\cite{Minard2016MEANTIMECorpus} and causality relations in the TCR corpus~\cite{Ning2018JointRelations}. We also intend to extend the list of supported corpora and baseline models, in particular, to support corpora that cast the TIE task as a question-answer problem, such as MCTaco~\cite{Zhou2019GoingUnderstanding} and TORQUE~\cite{Ning2020TORQUE:Questions}. This will allow us to produce a reproducibility study to investigate several state-of-the-art systems and benchmark them in the different corpora.

\begin{acks}
	We would like to acknowledge the financial support received during the development of this project. Hugo Sousa was financed by by National Funds through the FCT - Fundação para a Ciência e a Tecnologia, I.P. (Portuguese Foundation for Science and Technology) within the project StorySense, with reference 2022.09312.PTDC. Ricardo Campos and Alípio Jorge were financed by the ERDF European Regional Development Fund through the North Portugal Regional Operational Programme (NORTE 2020), under the PORTUGAL 2020 and by National Funds through the Portuguese funding agency, FCT - Fundação para a Ciência e a Tecnologia within project PTDC/CCI-COM/31857/2017 (NORTE-01-0145-FEDER-03185).
\end{acks}

\balance
\bibliographystyle{ACM-Reference-Format}
\bibliography{references.bib}

\end{document}

%% file: tables/data.tex
\begin{tabular*}{\textwidth}{c@{\extracolsep{\fill}}crrrr}
    \toprule
    & \textbf{Language} & \multicolumn{1}{c}{\textbf{\#~Docs}} & \multicolumn{1}{c}{\textbf{\#~Events}} & \multicolumn{1}{c}{\textbf{\#~Timexs}} & \multicolumn{1}{c}{\textbf{\#~Tlinks}} \\
    \midrule
    \multirow{8}{*}{AncientTimes \cite{Strotgen2014ExtendingDates}}                           & Arabic            & 5                                    & 0                                      & 106                                    & 0                                      \\
    & Dutch             & 5                                    & 0                                      & 130                                    & 0                                      \\
    & English           & 5                                    & 0                                      & 311                                    & 0                                      \\
    & French            & 5                                    & 0                                      & 290                                    & 0                                      \\
    & German            & 5                                    & 0                                      & 196                                    & 0                                      \\
    & Italian           & 5                                    & 0                                      & 234                                    & 0                                      \\
    & Spanish           & 5                                    & 0                                      & 217                                    & 0                                      \\
    & Vietnamese        & 4                                    & 0                                      & 120                                    & 0                                      \\
    \midrule
    Aquaint    \cite{UzZaman2013SemEval-2013Relations}                                                           & English           & 72                                   & 4,351                                  & 639                                    & 5,832                                  \\
    \midrule
    EventTime                 \cite{Reimers2016TemporalCorpus}                                  & English           & 36                                   & 1,498                                  & 0                                      & 0                                      \\
    \midrule
    GraphEVE   \cite{Glavas2015ConstructionGraphs}                                                                & English           & 103                                  & 4,298                                  & 0                                      & 18,204                                 \\
    \midrule
    KRAUTS  \cite{Strotgen2018KRAUTS:Corpus}                                                    & German            & 192                                  & 0                                      & 1,282                                  & 0                                      \\
    \midrule
    MATRES       \cite{Ning2018ARelations}                                                            & English           & 274                                  & 6,065                                  & 0                                      & 13,504                                 \\
    \midrule
    \multirow{4}{*}{MeanTime \cite{Minard2016MEANTIMECorpus}}                                             & English           & 120                                  & 1,882                                  & 349                                    & 1,753                                  \\
    & Spanish           & 120                                  & 2,000                                  & 344                                    & 1,975                                  \\
    & Dutch             & 120                                  & 1,346                                  & 346                                    & 1,487                                  \\
    & Italian           & 120                                  & 1,980                                  & 338                                    & 1,675                                  \\
    \midrule
    Narrative Container                    \cite{Bracchi2016EnrichringContainers}                               & Italian           & 63                                   & 3,559                                  & 439                                    & 737                                    \\
    \midrule
    \multirow{6}{*}{Professor Heideltime \cite{Sousa2023TEI2GO:Identification}}                                                 & English           & 24,642                               & 0                                      & 254,803                                & 0                                      \\
    & French            & 27,154                               & 0                                      & 83,431                                 & 0                                      \\
    & German            & 19,095                               & 0                                      & 194,043                                & 0                                      \\
    & Italian           & 9,619                                & 0                                      & 58,823                                 & 0                                      \\
    & Portuguese        & 24,293                               & 0                                      & 111,810                                & 0                                      \\
    & Spanish           & 33,266                               & 0                                      & 348,011                                & 0                                      \\
    \midrule
    Platinum \cite{UzZaman2013SemEval-2013Relations}                                                             & English           & 20                                   & 748                                    & 158                                    & 929                                    \\
    \midrule
    \multirow{4}{*}{TimeBank \cite{GuerreroNieto2011ModeSCorpus,Bittar2011FrenchCorpus,Costa2012TimeBankPT:Portuguese,UzZaman2013SemEval-2013Relations}} & Spanish           & 210                                  & 12,384                                 & 1,532                                  & 21,107                                 \\
    & French            & 108                                  & 2,115                                  & 533                                    & 2,303                                  \\
    & Portuguese        & 182                                  & 7,887                                  & 1,409                                  & 6,538                                  \\
    & English           & 183                                  & 6,681                                  & 1,426                                  & 5,120                                  \\
    \midrule
    TimeBank 1.2 \cite{UzZaman2013SemEval-2013Relations}                                                         & English           & 183                                  & 7,940                                  & 1,414                                  & 6,413                                  \\
    \midrule
    TCR       \cite{Ning2018JointRelations}                                                        & English           & 25                                   & 1,134                                  & 242                                    & 3,515                                  \\
    \midrule
    TDDiscourse \cite{Naik2019TDDiscourse:Events}                                                        & English           & 34                                   & 1,101                                  & 0                                      & 6,150                                  \\
    \midrule
    \multirow{7}{*}{TempEval 2 \cite{Verhagen2010SemEval-2010TempEval-2}}                                          & Chinese           & 52                                   & 4,783                                  & 946                                    & 7,802                                  \\
    & English           & 182                                  & 6,656                                  & 1,390                                  & 5,945                                  \\
    & French            & 83                                   & 1,301                                  & 367                                    & 372                                    \\
    & Italian           & 64                                   & 5,377                                  & 653                                    & 6,884                                  \\
    & Korean            & 18                                   & 2,583                                  & 317                                    & 0                                      \\
    & Spanish           & 210                                  & 12,384                                 & 1,502                                  & 13,304                                 \\
    & English           & 275                                  & 11,780                                 & 2,223                                  & 11,881                                 \\
    \midrule
    TimeBank Dense   \cite{Cassidy2014AnOrdering}                                                & English           & 36                                   & 1,712                                  & 289                                    & 12,715                                 \\
    \midrule
    TrainT3 \cite{UzZaman2013SemEval-2013Relations}                                                              & Spanish           & 175                                  & 10,686                                 & 1,269                                  & 17,283                                 \\
    \midrule
    \multirow{2}{*}{Wikiwars \cite{Mazur2010WikiWars:Expressions}}                                             & English           & 22                                   & 0                                      & 2,662                                  & 0                                      \\
    & German            & 22                                   & 0                                      & 2,239                                  & 0                                      \\
    \bottomrule
\end{tabular*}

%% file: tables/results.tex
\begin{tabular*}{\textwidth}{l@{\extracolsep{\fill}}ccccccccc} 
\hline
              & \multicolumn{3}{c}{\textbf{Platinum}} & \multicolumn{3}{c}{\textbf{TCR}} & \multicolumn{3}{c}{\textbf{MeanTime}}  \\ 
\cline{2-10}
              & $P$  & $R$  & $F_1$~($TF_1$)          & $P$  & $R$  & $F_1$~($TF_1$)     & $P$  & $R$  & $F_1$~($TF_1$)           \\ 
\hline
TimexBaseline & 88.1 & 75.4 & 81.2                    & 75.4 & 82.0 & 78.6               & 23.7 & 57.1 & 33.5                     \\
HeidelTime    & 84.0 & 79.4 & 81.8                    & 70.6 & 80.6 & 75.3               & 26.5 & 65.8 & 37.8                     \\
EventBaseline & 74.6 & 80.5 & 77.5                    & 48.3 & 92.6 & 63.5               & 25.8 & 54.1 & 34.9                     \\
CogCompTime2  & 39.7 & 39.7 & 39.7~(39.3)             & 75.4 & 75.4 & 75.4~(69.3)        & 30.7 & 28.6 & 29.6~(28.9)              \\
\hline
\end{tabular*}